# Increasing transformer token length with a Maximum Entropy Principle Method

R. I. Cukier

*Department of Chemistry, Michigan State University, East Lansing, Michigan 48824-1322, USA*

cukier@chemistry.msu.edu

Abstract

Transformers suffer from the computational overhead of their quadratic dependence on the length of sequences processed. We present three methods, all adding an intermediate step between training and inference/generation, which extend the autoregressive length of transformers. All rely on a Maximum Entropy Principle (MEP) whereby entropy is maximized in the presence of suitable constraints, accounted for by use of Lagrange Multipliers. These constraint methods extend the autoregressive character from $T$ to $2T$ tokens in a linear-with-$T$ fashion. There is overhead associated with this added step, but they should still be faster than the standard methods.

**1 Introduction**

Transformers have transformed the NLP world to a remarkable extent. Previous methodologies such as Recursive Neural Networks [Niu, Zaheer] suffer from the serial nature of that class of algorithms. The introduction of the attention concept with its intrinsic parallelization capabilities and its autoregressive nature to account for a history of tokens in predicting succeeding tokens is responsible for this explosion in NLP capabilities. [Vaswani, Phuong] However, as apparent from the autoregressive nature of attention, for $T$ context tokens the scaling is $O(T^2)$, limiting the accounted-for history. Schemes to alleviate this problem are available, concentrated on various masking/sparsity schemes as implemented in both algorithm and hardware. [Child, Beltagy, Devlin, Zaheer, Jiang] There still are other aspects of transformers that warrant attention. [Dong]

This work proposes three methods to increase the considered token length. They are directed to decoder-only transformers [Phuong] that use a trained model for inference/generation. All add an intermediate step between training and inference/generation. They are based on a Maximum Entropy Principle (MEP), introduced to Statistical Mechanics by Jaynes [Jaynes], whereby entropy is maximized in the presence of suitable constraints on the various probabilities, as enforced by Lagrange Multipliers. This added calculation has an associated overhead from the use of some nonlinear optimization scheme. But, for long token ranges, methods that extend from $T$ to $2T$ tokens in a linear-with-$T$ approach may still be faster than training with backpropagation algorithms to determine a *very* large number of parameters.

An earlier study [Lou] showed that a Markov *chain* (versus process) could be augmented using a MEP method. Schematically: $p(3|2), p(2|1), p(3|1) \xrightarrow{MEP} p(3|21)$. Neighbor, $p(3|2)$ and $p(2|1)$, and next neighbor, $p(3|1)$, conditional probabilities could accurately produce the "augmented" conditional $p(3|21)$. In the language of autoregressive LLMs: $p(w_t | w_{t-1}), p(w_{t-1} | w_{t-2}), p(w_t | w_{t-2}) \rightarrow p(w_t | w_{t-1} w_{t-2})$.

This work extends the MEP to estimate

$$P(w_{1:T} | W_0) = \prod_{t=1}^{T} P(w_t | w_{1:t-1}, W_0), \text{ with } w_{1:0} = 0, W_0 := 0. \tag{1.1}$$

The individual terms in Eq (1.1) are to be obtained using some attention method so that

$$P(w_t | w_{1:t-1}, W_0) \Leftarrow \text{attn}(Q | \mathbf{K}); \text{ with a } Q(\text{uery}) \text{ and } \mathbf{K}(\text{ey}) \text{ denoting the } 1:t-1 \text{ keys.} \tag{1.2}$$

It will be convenient in the following to use integer notation for tokens, and it matters not if one thinks of going forward or backward in tokens. Then, with "MC" for Markov chain, "AMC" for augmented Markov chain and AUTO for Eq (1.1)

$$\begin{aligned} P_{MC}(w_{1:T} | W_0 = 0) &= [P(1|0) P(2|1)] P(3|2) P(4|3) P(5|4) \ldots \\ P_{AMC}(w_{1:T} | W_0 = 0) &= [P(1|0) P(2|1)] P(3|21) P(4|32) P(5|43) \ldots \\ &\cdots \\ P_{AUTO}(w_{1:T} | W_0 = 0) &= [P(1|0) P(2|10)] P(3|210) P(4|3210) P(5|43210) \ldots \end{aligned} \tag{1.3}$$

It is worth pointing out, as previously shown [Lou], that among all possible joint distributions the Markov chain $P_{MC}$ in Eq. (1.3) maximizes entropy when the neighbor pair probabilities are restricted to their estimated values. Thus, if only neighbor pair probabilities are specified, the Markov chain is the *optimal* solution. The same holds true for $P_{AMC}$ in Eq.(1.3). And it holds for the higher-order chains by induction.

In the following we introduce other token sets that could be of interest. For example, see Figure 1 in [Zaheer] where various "masking/chunking" schemes of non-contiguous tokens are proposed. They all are attempts to reduce the $O(T^2)$ scaling.

## 2.1 Method 1 $\mathrm{MEP}[T]$

This method generalizes our Maximum Entropy Principle, $\mathrm{MEP} := \mathrm{MEP}[1]$, to higher-order token groupings. It is a masking/chunking method as the tokens used for construction of the desired probabilities are not necessarily contiguous.

$\mathrm{MEP}[1]$: $p(1|2); p(2|3); p(1|3) \to p(1|23);$
$\mathrm{MEP}[2]$: $p(1|23); p(2|45); p(1|45) \to p(1|2345);$
$\mathrm{MEP}[3]$: $p(1|234); p(2|567); p(1|567) \to p(1|234567);$ (2.1.1)
$\ldots\ldots\ldots\ldots$
$\mathrm{MEP}[T]$: $p(1|g_1); p(2|g_2); p(1|g_2) \to p(1|g_1,g_2); \; T := \dim(g_1 = g_2)$

where, for $MEP[T]$, introduce the notation and definitions:

$$g_1 := (2,3,4,\ldots,T+1); \; g_1^+ := (3,4,\ldots,T+1); \; g_2 := (T+2, T+3, \ldots, 2T+1); \; g := (g_1, g_2). \quad (2.1.2)$$

The optimized probability is obtained from:

$$f^*(1, g_1, g_2) = \arg\max_f \left\{ -\sum_{1, g_1, g_2} f(1, g_1, g_2) \ln f(1, g_1, g_2) \right\}$$

subject to the 3 constraint equations:

$$\sum_1 \sum_{g_1^+} f(1, g_1, g_2) - p(2, g_2) = 0; \quad (2.1.3)$$

$$\sum_{g_2} f(1, g_1, g_2) - p(1, g_1) = 0;$$

$$\sum_{g_1} f(1, g_1, g_2) - p(1, g_2) = 0.$$

The associated Lagrangian is:

$$L := -\sum_{1,g_1,g_2} f(1,g_1,g_2) \ln f(1,g_1,g_2) + \sum_{2,g_2} \lambda^{2,g_2} \left[ \sum_{1} \sum_{g_1^+} f(1,g_1,g_2) - p(2,g_2) \right]$$
$$+ \sum_{1,g_1} \lambda^{1,g_1} \left[ \sum_{g_2} f(1,g_1,g_2) - p(1,g_1) \right] + \sum_{1,g_2} \lambda^{1,g_2} \left[ \sum_{g_1} f(1,g_1,g_2) - p(1,g_2) \right] \quad (2.1.4)$$

(These expressions indicate sums over random variable (RV) *elements* using the RV *names*, to simplify notation).

There are $2T+1$ RVs that are split into $T+1$ Lagrange Multipliers (LMs) and $T$ other RVs in $f(1,g_1,g_2)$. E.g., for $T=3$ there are 4 LMs and 3 other RVs summed in the brackets in Eq. (2.1.4). The resulting Euler-Lagrange equation is

$$-1 - \ln f(1,g_1,g_2) + \lambda^{2,g_2}_{i_2,i_{g_2}} + \lambda^{1,g_1}_{i_1,i_{g_1}} + \lambda^{1,g_2}_{i_1,i_{g_2}} = 0; \quad i_1, i_2, i_{g_1}, i_{g_2} \in \mathcal{I} \quad (2.1.5)$$

where we use $i_{g_2}$ for elements of RVs $g_2$, etc.

$MEP[G>1]$ does have an "asymmetry" issue. That is, $g_1$ is a "neighbor set" of RV 1, but the $g_2$ set is removed in token-ordered space from RV 1, as displayed in Eq. (2.1.1).

For this $T=3$ example: $g_1 = (2,3,4)$, $g_1^+ = (3,4)$, $g_2 = (5,6,7)$, $g = (2,3,4,5,6,7)$. The constraint equations are:

$$\sum_{2,5,6,7} \lambda^{2,5,6,7} \left[ \sum_{1} \sum_{3,4} f(1,2,3,4,5,6,7) - p(2,5,6,7) \right] = 0;$$
$$\sum_{1,2,3,4} \lambda^{1,2,3,4} \left[ \sum_{5,6,7} f(1,2,3,4,5,6,7) - p(1,2,3,4) \right] = 0; \quad (2.1.6)$$
$$\sum_{1,5,6,7} \lambda^{1,5,6,7} \left[ \sum_{2,3,4} f(1,2,3,4,5,6,7) - p(1,5,6,7) \right] = 0.$$

The $p(Q|K)$ terms: $p(1,g_1)$, $p(1,g_2)$ and $p(2,g_2)$ provide the optimized $f^*(1,g_1,g_2)$. As noted in the Introduction, there is a trade-off between training and this intermediate MEP step. Thus, the scaling with the number of tokens, $T$, and the number of RV elements, $\mathcal{I}$, is crucial. There are $T+1$ LMs and 3 constraint equations with scaling $\sim 3 * \mathcal{I}^{T+1}$ that is ostensibly unfavorable. However, $\mathcal{I}$, the number of terms to sum over in the three probabilities in the Euler-Lagrange equations is limited to mainly $\mathcal{I}=1$, for normal operation. That will almost always be the case unless the generative "temperature" is set quite high. That $\mathcal{I}=1$ is the most likely case will be supported by a calculation provided in **Appendix A**. There, it is shown that as the key length increases, other things being equal, the largest value of $p(Q|\mathbf{K}) \to 1$. (More precisely, we show that $|p_{\max}(Q|\mathbf{K}) - p_{\min}(Q|\mathbf{K})| \to 1$). Note that a virtue of $MEP[T]$ is a "doubling" of the autoregression length.

There are redundancies in the RV element sums in the equations of constraint that need to be eliminated in the method used to determine their values. That technical issue is addressed in **Appendix C**.

### 2.2 Method 2: GMEP (Generalized MEP → GMEP)

Here, the idea is to increase the number of two-point conditional probabilities to constrain and use them to do a MEP to produce an autoregressive form: E.g. $p(1|2), p(1|3), p(1|4); p(2|3), p(2|4); p(3|4) \to p(1234)$.

This pair token approach is a generalization of the MEP[1] scheme in Eq. (2.1.1). For the above, there are six constraint equations. Schematically:

$$\sum_{34} f(1234) - p(12) = \sum_{24} f(1234) - p(13) = \sum_{23} f(1234) - p(14)$$

$$= \sum_{14} f(1234) - p(23) = \sum_{13} f(1234) - p(24) = \sum_{12} f(1234) - p(34) = 0.$$

In general, there are $T$ RVs determined by the GMEP, where:

RV 1 is determined by $T-1$ RVs: $2, 3, 4..., T$;

RV 2 is determined by $T-2$ RVs: $3, 4, 5..., T$;

……………………………………………………………..

RV $T-1$ is determined by 1 RV: $T$.

The number of constraint equations $S_T$ is

$$S_T = (T-1) + (T-2) + ... + 2 + 1 = (T-1)T/2. \tag{2.2.1}$$

The generalization of the above to the $(T-1)T/2$ constraints equations is

$$\overset{*}{\sum} f(12...T) - p(1,2) = \overset{*}{\sum} f(12...T) - p(1,3) = ... = \overset{*}{\sum} f(12...T) - p(T-1,T) = 0, \quad (2.2.2)$$

where the * denotes exclude from the sums the RVs in the corresponding fixed probabilities.

For the token set $\mathcal{T} = \{t_1, t_2, ..., t_T\} \to \{123..t..T\}$ define

$$\mathcal{T}_- = \mathcal{T} - \{t, t' \mid t < t'\} \quad (2.2.3.\text{A})$$

with corresponding RV elements

$$\mathcal{I}_- = \mathcal{I} - \{i_t, i_{t'} \mid t < t'\}. \quad (2.2.3.\text{B})$$

For simplicity, assume that all RV elements are in the same alphabet $i_t \in \mathcal{X}_\mathcal{I}$

The constraint equation for token pair $\{t, t'\}$ with $(1 \le t < t' \le T)$ is

$$\sum_{\mathcal{I}_-} f_\mathcal{T}(\mathcal{I}) - p_{t,t'}(i_t, i_{t'}) = 0 \quad (1 \le t < t' \le T). \quad (2.2.4)$$

The Lagrangian is

$$L = -\sum_\mathcal{I} f_\mathcal{T}(\mathcal{I}) \ln f_\mathcal{T}(\mathcal{I})$$
$$+ \sum_{i_1, i_2} \lambda^{1,2}_{i_1, i_2} \left[ \sum_{\mathcal{I}_-} f_\mathcal{T}(\mathcal{I}) - p_{1,2}(i_1, i_2) \right] + \sum_{i_1, i_3} \lambda^{1,3}_{i_1, i_3} \left[ \sum_{\mathcal{I}_-} f_\mathcal{T}(\mathcal{I}) - p_{1,3}(i_1, i_3) \right] \quad (2.2.5)$$
$$+ ... + \sum_{i_{T-1}, i_T} \lambda^{T-1,T}_{i_{T-1}, i_T} \left[ \sum_{\mathcal{I}_-} f_\mathcal{T}(\mathcal{I}) - p_{T-1,T}(i_{T-1}, i_T) \right].$$

The resulting Euler-Lagrange equation is

$$-1 - \ln f_\mathcal{T}(\mathcal{I}) + \lambda^{1,2}_{i_1, i_2} + \lambda^{1,3}_{i_1, i_3} + ... + \lambda^{T-1,T}_{i_{T-1}, i_T} = 0. \quad (2.2.6)$$

Noting that

$$S := \sum_{j=1}^{j=T-1} \sum_{k=j+1}^{T} \lambda^{j,k}_{i_j, i_k} \quad (2.2.7)$$

will generate the summed LM terms above, the Euler-Lagrange equation can be compacted to

$$-1 - \ln f_\mathcal{T}(\mathcal{I}) + S = 0. \quad (2.2.8)$$

The GMEP, defined by equations (2.2.6-8), uses all (trained) pair probabilities for token pairs $t$ and $t'$ s.t. $1 \le t < t' \le T$. They are used to construct the $p_{12...T}(i_1, i_2, ..., i_T)$ for the inference/generative phase. The scaling follows from the $(T-1)T/2 * \mathcal{I}^2 \sim T^2/2 * \mathcal{I}^2$ LMs to be determined. Again, there are redundancies to eliminate, as shown in **Appendix C**.

For the GMEP, the *training* phase is much cheaper because the autoregression is kept to only one token back, thus linear vs. quadratic in tokens. That is at the expense of, in the MEP calculation, determining the many Lagrange Multipliers in Eq. (2.2.7), though.

**2.3 Method 3: SMEP (Symmetrized MEP)**

The SMEP idea is to make a "symmetrized" version of the original MEP, and extend it in both directions to additional tokens: $0|-1,-2,...,-T$ and $0|1,2,...,T$. The corresponding probabilities are $p(-T,-T+1,...,-1,0)$ and $p(0,1,...,T-1,T)$. This approach fits in with the 'mask' concept in standard Transformers.

Of the probabilities to constrain, two have $T+1$ RV terms: $p(0,1,...,T-1,T)$ and $p(-T,-T+1,...,-1,0)$. The third probability to constrain is $p(-T,-(T-1),...,-1,1,...,T-1,T)$. It has a two-times-larger number of terms to constrain then the other two constraints. To avoid this deficiency, and provide an even more symmetric formulation, split this third probability to constrain into $2T$ "neighbor"-type probabilities. E.g., for $T=2$, split the third probability into 4 triplet probabilities to constrain:

$$p(-2,-1,+1,+2) \to \{p(-2,1,2), p(-1,1,2), p(1,-1,-2), p(2,-1,-2)\}.$$

In general, there are $2T+1$ RVs and $2+2T$ constraint equations: 2 from the $p(-T,-T+1,...,1,0)$ and $p(0,1,...,T-1,T)$ probabilities and $2T$ from the split of the order $2T$ third probability, $p(-T,-(T-1),...,-1,1,...T-1,T)$, into $2T$ terms, each also with $T+1$ RVs. Thus, the probabilities to constrain are written schematically as follows:

$$\begin{bmatrix} p(-(T),1,2,...,T), \\ p(-(T-1),1,2,...,T), \\ .... \\ p(-(1),1,2,...,T) \end{bmatrix} \quad (T \text{ split from nonstandard constraint})$$

$$[p((0),1,2,...,T)] \quad \text{(one from first constraint)}$$

$$[p((0),-1,-2,...,-T)] \quad \text{(one from second constraint)} \quad (2.3.1)$$

$$\begin{bmatrix} p((1),-1,-2,...,-T), \\ ... \\ p((T-1),-1,-2,...,-T), \\ p((T),-1,-2,...,-T) \end{bmatrix} \quad (T \text{ split from nonstandard constraint})$$

As an example, for $T = 3$ with 7 RVs there are 8 constraint equations, each with 4 RVs.

$$\begin{bmatrix} \sum^{**} \lambda^{**} \left[ \sum^{*} f(-3,-2,-1,0,1,2,3) - p(-(3),1,2,3) \right] = 0 \\ \sum^{**} \lambda^{**} \left[ \sum^{*} f(-3,-2,-1,0,1,2,3) - p(-(2),1,2,3) \right] = 0 \\ \sum^{**} \lambda^{**} \left[ \sum^{*} f(-3,-2,-1,0,1,2,3) - p(-(1),1,2,3) \right] = 0 \\ \sum^{**} \lambda^{**} \left[ \sum^{*} f(-3,-2,-1,0,1,2,3) - p(-(0),1,2,3) \right] = 0 \end{bmatrix} \quad (2.3.2.\text{A})$$

$$\begin{bmatrix} \sum^{**} \lambda^{**} \left[ \sum^{*} f(-3,-2,-1,0,1,2,3) - p((0),-1,-2,-3) \right] = 0 \\ \sum^{**} \lambda^{**} \left[ \sum^{*} f(-3,-2,-1,0,1,2,3) - p((1),-1,-2,-3) \right] = 0 \\ \sum^{**} \lambda^{**} \left[ \sum^{*} f(-3,-2,-1,0,1,2,3) - p((2),-1,-2,-3) \right] = 0 \\ \sum^{**} \lambda^{**} \left[ \sum^{*} f(-3,-2,-1,0,1,2,3) - p((3),-1,-2,-3) \right] = 0 \end{bmatrix} \quad (2.3.2.\text{B})$$

*/** Indicates a sum over the elements of the 3 RVs *not appearing* /4 RVs *appearing* in the probabilities that are to be constrained. E.g., the first term in (2.3.2.A) is

$$\sum_{-3,1,2,3} \lambda^{-3,1,2,3} \left[ \sum_{-2,-1,0} f(-3,-2,-1,0,1,2,3) - p(-(3),1,2,3) \right] = 0. \tag{2.3.3}$$

Once the third constraint is split into $2T$ terms, what appears in this SMEP scheme is a split between positive and negative token sides (so symmetric) with the constraints in turn on RVs from $0$ to $T$ in the presence of all the other side RVs: $-1, -2, \ldots -T$, and *vice versa*.

To write the SMEP compactly, introduce $t_{\mathcal{T}} := (t_{-T}, t_{-T+1}, \ldots, t_0, \ldots, t_{T-1}, t_T)$. The entropy (using RV names instead of RV elements) is

$$H = -\sum_{t_{-T}} \sum_{t_{-T+1}} \cdots \sum_{t_0} \cdots \sum_{t_{T-1}} \sum_{t_T} f(t_{-T}, t_{-T+1}, \ldots, t_0, \ldots, t_{T-1}, t_T) \ln f(t_{-T}, t_{-T+1}, \ldots, t_0, \ldots, t_{T-1}, t_T)$$

$$= -\prod_{t=t_{-T}}^{t=t_{+T}} \sum_{t \in \mathcal{X}} f(t_{\mathcal{T}}) \ln f(t_{\mathcal{T}}) \text{ with } t_{\mathcal{T}} := (t_{-T}, t_{-T+1}, \ldots, t_0, \ldots t_{T-1}, t_T) \tag{2.3.4}$$

Introduce some definitions to separate the $2+2T$ constraint equations into the $T+1$ LM terms and the $T$ interior sum terms :

$$t_{\mathcal{T}_{-g}^+} := (-t_g, t_1, t_2, \ldots, t_T) \ (T+1 \text{ terms}); \ t_{\mathcal{T}^-} := (t_{-T+1}, t_{-T+2}, \ldots, t_{-1}, t_0) \ (T \text{ terms})$$

$$t_{\mathcal{T}_{+g}^-} := (+t_g, t_{-1}, t_{-2}, \ldots, t_{-T}) \ (T+1 \text{ terms}) \ t_{\mathcal{T}^+} := (t_0, t_1, \ldots, t_{T-2}, t_{T-1}) \ (T \text{ terms}) \tag{2.3.5}$$

$$t_g \in (0, 1 \ldots, T)$$

The constrained optimization is

$$f^*(t_{\mathcal{T}}) = \overset{\arg\max}{f} \left\{ -\prod_{t=-t_T}^{t=+t_T} \sum_{t \in \mathcal{X}} f(t_{\mathcal{T}}) \ln f(t_{\mathcal{T}}) \right\}$$

subject to the $2T+2$ constraint equations:

$$\sum_{t_{\mathcal{T}^-}} f\left(t_{\mathcal{T}_{-g}^+}, t_{\mathcal{T}^-}\right) - p\left(t_{\mathcal{T}_{-g}^+}\right) = 0 \ (T+1 \text{ terms}, t = 0, 1, \ldots, T) \tag{2.3.6}$$

$$\sum_{t_{\mathcal{T}^+}} f\left(t_{\mathcal{T}_{+g}^-}, t_{\mathcal{T}^+}\right) - p\left(t_{\mathcal{T}_{+g}^-}\right) = 0 \ (T+1 \text{ terms}, -t = 0, 1, \ldots, T)$$

Thus, the function to MEP, $f(-T,-(T-1),\ldots -1,0,1,\ldots,T-1,T)$, has to be summed over the $T$ RVs not appearing in the $T+1$ probabilities that are to be constrained with LMs for the total of $2T+1$ RVs under consideration. The corresponding Lagrangian is

$$L = -\prod_{t=-t_T}^{t=+t_T}\sum_{t\in\mathcal{X}} f(t_\mathcal{T})\ln f(t_\mathcal{T})$$

$$+\sum_{t_{\mathcal{T}_{-g}^+}} \lambda^{t_{\mathcal{T}_{-g}^+}}\left[\sum_{t_{g^-}} f(t_{\mathcal{T}_{-g}^+}, t_{\mathcal{T}^-}) - p(t_{\mathcal{T}_{-g}^+})\right] + \sum_{t_{\mathcal{T}_{+g}^-}} \lambda^{t_{\mathcal{T}_{+g}^-}}\left[\sum_{t_{g^+}} f(t_{\mathcal{T}_{+g}^-}, t_{\mathcal{T}^+}) - p(t_{\mathcal{T}_{+g}^-})\right].$$

(2.3.7)

The resulting Euler-Lagrange equation is

$$-1 - \ln f(t_\mathcal{T}) + \sum_{t_{\mathcal{T}_{-g}^+}} \lambda^{t_{\mathcal{T}_{-g}^+}} + \sum_{t_{\mathcal{T}_{+g}^-}} \lambda^{t_{\mathcal{T}_{+g}^-}} = 0.$$

(2.3.8)

In the SMEP, the number of constraint equations scales linearly with $2T+1$, the number of autoregressive tokens considered. There are $2T+2$ constraint equations, each with $T+1$ LM RVs to sum over. Thus, there are $\mathcal{I}^{T+1}$ LMs to determine in each constraint equation and the scaling is $\sim (2T+2)\mathcal{I}^{T+1}$.

What is the advantage of the SMEP? From a $T+1$ token trained set, SMEP bootstraps to a $2T+1$ token set. As training $\sim T^2$, the task is reduced to order $T$, with the overhead of evaluation of $\sim 2T$ LM constraints.

## 3 Conclusion

The MEP approaches presented here are designed to increase the number of considered tokens for inference/generation appropriate to decoder-only transformers. They do have the tradeoff of requiring an intermediate step between training and inference/generation with an associated overhead, relative to increasing the training length with its $T^2$ character. Optimizing a set of Lagrange Multipliers should be much less compute-intensive than that required for training by backpropagation for a very large set of parameters. Our previous work [Lou] used a Newton-Krylov method for the nonlinear optimization required to determine the Lagrange

Multipliers. It always converged to numerical tolerance, and produced probabilities with close to the exact (but larger) entropy.

The MEP scalings are: $\text{MEP}[T] \sim 3\mathcal{I}^{T+1}$, $\text{GMEP} \sim (T^2/2)\mathcal{I}^2$, and $\text{SMEP} \sim (2T+2)\mathcal{I}^{T+1}$ for $T$ tokens and $\mathcal{I}$ RV elements of significant probability. Should $\mathcal{I}$ turn out to be large, then both $\text{MEP}[T]$ and SMEP may become impractical. However, as demonstrated in **Appendix A**, the longer the context considered, the greater the difference between $p_{\max}$ and $p_{\min}$, indicating that $\mathcal{I} \to 1$. Then, these methods should be practical. Note that the **Appendix A** proof shows that increasing the number of variables conditioned on increases the difference between smallest and largest probabilities of the desired random variable *term-by-term,* versus for the corresponding entropies that are averaged quantities.

Acknowledgement. We are in debt to Dr. Hongfeng Lou for the elegant proof of limits on conditional probability values presented in **Appendix A**.

**Appendix A**

Two sequences of inequalities are proved in this Appendix; the first on conditional entropy and the second on conditional probabilities. Our considerations for the entropy rely on Cover and Thomas' (C-T) book [Cover]. For probabilities, we aren't aware of the derivation provided below.

An $N$ random variable probability mass function $p(x_1,..,x_N)$ can be decomposed into a product of conditionals $p(x_i | x_{i-1},...,x_1)$ using a "chain rule" as follows:

$$p(x_1,..,x_N) = \prod_{i=1}^{N} p(x_i | x_{i-1},...,x_1). \tag{A.1}$$

This decomposition is well-suited to an analysis of autoregressive processes. The corresponding entropy $H(x_1,..,x_N)$ can also be decomposed in analogous, but additive, fashion:

$$H(x_1,..,x_N) = \sum_{i=1}^{N} H(x_i | x_{i-1},...,x_1). \tag{A.2}$$

These expressions lead to the explicit autoregressive forms for probabilities:

$$p_{IND} := p^{(1)}(x_1,..,x_N) = p(x_N)...p(x_2)p(x_1)$$

$$p_{MC} := p^{(2)}(x_1,..,x_N) = p(x_N | x_{N-1})...p(x_2 | x_1)p(x_1)$$

$$p_{AMC} := p^{(3)}(x_1,..,x_N) = p(x_N | x_{N-1}, x_{N-2})...p(x_2 | x_1)p(x_1) \quad (A.3)$$

........

$$p^{(n)}(x_1,..,x_N) = \prod_{i=1}^{N} p(x_i | x_{i-1},...,x_{i-n}); (i-n) > 0.$$

Above, $p_{IND}$ denotes independent, $p_{MC}$ Markov and $p_{AMC}$ augmented Markov that was pursued for the MEP approach [Lou]. The corresponding entropies are

$$H_{IND}(x_1,..,x_N) := H^{(1)} = H(x_1) + H(x_2) + \sum_{i=3}^{N} H(x_i)$$

$$H_{MC}(x_1,..,x_N) := H^{(2)} = H(x_1) + H(x_2 | x_1) + \sum_{i=3}^{N} H(x_i | x_{i-1})$$

$$H_{AMC}(x_1,..,x_N) := H^{(3)} = H(x_1) + H(x_2 | x_1) + \sum_{i=3}^{N} H(x_i | x_{i-1}, x_{i-2}) \quad (A.4)$$

.......

$$H^{(n)} = \sum_{i=1}^{N} H(x_i | x_{i-1},...,x_{i-n}); (i-n) > 0.$$

The MEP work obtained a generalization of the fundamental result

$$H(X | Y) \leq H(X) \quad (A.5)$$

whereby conditioning reduces entropy; or, as said evocatively in C-T, "Information can't hurt." Actually, for our purposes, information can help, as shown below for conditional probabilities. We generalized Eq. (A.5) to

$$H(x_i | x_{i-1}, x_{i-2}) \leq H(x_i | x_{i-1}) \leq H(x_i). \quad (A.6)$$

It is straightforward to further generalize this to

$$H(x_i | x_{i-1}, x_{i-2},...,x_1) \leq H(x_i | x_{i-1}, x_{i-2},...,x_2) \leq \cdots \leq H(x_i | x_{i-1}) \leq H(x_i). \quad (A.7)$$

The above entropy results are known, or are generalizations of known principles. They follow from use of Jensen's inequality on the indicated conditional entropy [Cover] that is the expectation of the logarithm of the corresponding conditional probability, noting that the logarithm is a strictly concave function.

Turning now to inequalities on conditional probabilities, the goal is to show that as more random variables (RVs) are fixed, the spread in probability values increases. To that end, given three RVs: $X_1$, $X_2$, $X_3$, the following inequalities will be shown to hold:

$$\min_{x_3,x_2,x_1} p(x_3 | x_2, x_1) \leq \min_{x_3,x_2} p(x_3 | x_2) \leq \max_{x_3,x_2} p(x_3 | x_2) \leq \max_{x_3,x_2,x_1} p(x_3 | x_2, x_1). \quad (A.8)$$

Using the law of iterated expectation, note that

$$p(x_3 | x_2) = E_{X_3}\left[1_{X_3=x_3} | X_2 = x_2\right] = E_{X_1}\left[E_{X_3}\left[1_{X_3=x_3} | X_2 = x_2, X_1\right] | X_2 = x_2\right]. \quad (A.9)$$

Define

$$p(x_3 | x_2, X_1) := E_{X_3}\left[1_{X_3=x_3} | X_2 = x_2, X_1\right]. \quad (A.10)$$

(The expectation over indicator function $1_{X_3=x_3}$ pulls out the corresponding desired probability).

Therefore

$$p(x_3 | x_2) = E_{X_1}\left[p(x_3 | x_2, X_1) | X_2 = x_2\right]. \quad (A.11)$$

$p(x_3 | x_2)$ can be thought of as some average of $p(x_3 | x_2, X_1)$ over $X_1$, given fixed $x_2$ and $x_3$, implying that $p(x_3 | x_2)$ is less than the maximum and greater than the minimum of $p(x_3 | x_2, X_1)$ over $X_1$, given fixed $x_2$ and $x_3$.

The following proves the above intuition. Given any fixed $x_2$ and $x_3$

$$\min_s p(x_3 | x_2, X_1 = s) \leq p(x_3 | x_2, X_1) \leq \max_s p(x_3 | x_2, X_1 = s) \quad (A.12)$$

by definition of minimum and maximum. Therefore

$$\min_s p(x_3 | x_2, X_1 = s) \leq E_{X_1}\left[p(x_3 | x_2, X_1) | X_2 = x_2\right] \leq \max_s p(x_3 | x_2, X_1 = s) \quad (A.13)$$

and, by using Eq. (A.11),

$$\min_s p(x_3 | x_2, X_1 = s) \leq p(x_3 | x_2) \leq \max_s p(x_3 | x_2, X_1 = s). \quad (A.14)$$

Therefore

$$\min_{s,t,u} p(X_3 = u | X_2 = t, X_1 = s) \leq \min_s p(x_3 | x_2, X_1 = s) \leq p(x_3 | x_2) \quad (A.15)$$

with the first inequality above obtained by definition of minimum and

$$p(x_3 | x_2) \leq \max_s p(x_3 | x_2, X_1 = s) \leq \max_{s,t,u} p(X_3 = u | X_2 = t, X_1 = s) \quad (A.16)$$

with the second inequality above obtained by definition of maximum.

Note that the inequalities in Eq.( A.15) and (A.16) are first for $s$ and then for $s,t,u$, so $s$ is the same in both, and fixes $t,u$.

$$\therefore \min_{s,t,u} p(X_3 = u \mid X_2 = t, X_1 = s) \le p(x_3 \mid x_2) \le \max_{s,t,u} p(X_3 = u \mid X_2 = t, X_1 = s) \quad \square \qquad (A.17)$$

The $\min_{s,t,u} p(X_3 = u \mid X_2 = t, X_1 = s)$ is a constant and is a lower bound on $p(x_3 \mid x_2)$ and the $\max_{s,t,u} p(X_3 = u \mid X_2 = t, X_1 = s)$ is a constant and an upper bound on $p(x_3 \mid x_2)$.

Finally, then,

$$\begin{aligned}
\min_{s,t,u} p(X_3 = u \mid X_2 = t, X_1 = s) &\le \min_{u,t} p(X_3 = u \mid X_2 = t) \\
&\le \max_{u,t} p(X_3 = u \mid X_2 = t) \le \max_{s,t,u} p(X_3 = u \mid X_2 = t, X_1 = s) \quad \square
\end{aligned} \qquad (A.18)$$

Thus, the $X_1$ conditionals lower and upper bound the unconditional on $X_1$ probability. While the *entropy* decrease with conditioning in Eq. (A.7) is suggestive, it is on the average. Eq. (A.18) shows that an increase in conditioning increases the difference between smallest and largest probabilities of the desired random variable. That is the origin of the statement before Eq. (A.6) that, for our purposes, information can help in limiting $\mathcal{I}$, the number of RV elements of significant probability.

Straightforward generalization of the number of RVs from $X_2 \to X_{N-1}, X_{N-2}, ..., X_2$ provides

$$\begin{aligned}
&\min_{x_N, x_{N-1},...,x_2, s} p(X_N = x_N \mid X_{N-1} = x_{N-1}, X_{N-2} = x_{N-2}, ..., X_2 = x_2, X_1 = s) \\
&\le \min_{x_N, x_{N-1},...,x_2} p(X_N = x_N \mid X_{N-1} = x_{N-1}, X_{N-2} = x_{N-2}, ..., X_2 = x_2) \\
&.... \\
&\le \min_{x_N, x_{N-1}} p(X_N = x_N \mid X_{N-1} = x_{N-1}) \\
&\le \max_{x_N, x_{N-1}} p(X_N = x_N \mid X_{N-1} = x_{N-1}) \\
&.... \\
&\le \max_{x_N, x_{N-1},...,x_2} p(X_N = x_N \mid X_{N-1} = x_{N-1}, X_{N-2} = x_{N-2}, ..., X_2 = x_2) \\
&\le \max_{x_N, x_{N-1},...,x_2, s} p(X_N = x_N \mid X_{N-1} = x_{N-1}, X_{N-2} = x_{N-2}, ..., X_2 = x_2, X_1 = s).
\end{aligned} \qquad (A.19)$$

Furthermore, from Eq. (A.19) and from Eq. (A.3), which multiplies these conditional probabilities,

$$\min p^{(N-1)}(x_1,..,x_N) \leq ... \leq \min p^{(2)}(x_1,..,x_N)$$
$$\leq \max p^{(2)}(x_1,..,x_N) \leq ... \leq \max p^{(N-1)}(x_1,..,x_N) \tag{A.20}$$

Thus, in the context of autoregressive attention, the chance that more than one long-context probability is likely is significantly decreased. That may explain why it is often observed that a too-high "temperature" at inference/generation leads to hallucinatory behavior.

**Appendix B.**

Here, a model is introduced that illustrates that as $\Delta p \doteq p_{max} - p_{min}$ increases the corresponding entropy decreases. The MEP applied to a probability distribution defined on the set $\{x_1, x_2, ..., x_K\}$ with the constraint that the mean $\mu$ is known provides

$$p(X = x_k) = Cr^{x_k} \tag{B.1}$$

Setting $x_k = k$ gives $C = \dfrac{1}{\mu - 1}$ and $r = \dfrac{\mu - 1}{\mu}$, so

$$p_k := p(X = k) = \frac{1}{\mu - 1}\left(\frac{\mu - 1}{\mu}\right)^k \quad (1 \leq \mu \leq \infty). \tag{B.2}$$

As $\mu$ increases the probabilities become more uniform:

$$\frac{p_{k+1}}{p_k} = \left(\frac{\mu - 1}{\mu}\right) \quad (1 \leq \mu \leq \infty). \tag{B.3}$$

The entropy $H(\mu)$ can be obtained from the $p_k$ distribution in Eq. (B.3) with

$$\log p_k = \log\left(\frac{1}{\mu - 1}\right) + k \log\left(\frac{\mu - 1}{\mu}\right) = (k-1)\log(\mu - 1) - k \log \mu \tag{B.4}$$

as

$$H(\mu) = S_0 \log(\mu - 1) + S_1 \log\left[\frac{\mu}{\mu - 1}\right], \tag{B.5}$$

where

$$S_0 := \sum_{k=1}^{\infty} P_k = \text{norm} = \frac{1}{\mu - 1} := C; \quad S_1 := \sum_{k=1}^{\infty} k P_k = \mu. \tag{B.6}$$

Thus

$$H(\mu) = \log(\mu-1) + \mu \log\left[\frac{\mu}{\mu-1}\right] = \mu \log \mu - (\mu-1)\log(\mu-1). \tag{B.7}$$

The limiting behaviors are:

$$\lim_{\mu \to 0} H(\mu) = 0; \quad \lim_{\mu \to \infty} H(\mu) = 1. \tag{B.8}$$

The entropy monotonically increases from $0$ to $1$ as $\mu$ increases from $1$ to $\infty$:

$$0 \leq H(\mu) \leq 1; (1 \leq \mu \leq \infty); \quad H(\mu_2) \geq H(\mu_1) \text{ for } \mu_2 \geq \mu_1. \tag{B.9}$$

Now consider the max range issue $\Delta p \doteq p_{\max} - p_{\min}$:

$$\Delta p(\mu \to 1) = |1 - 0| = 1; \text{ where } H(\mu = 1) = 0 \tag{B.10}$$

and

$$\Delta p(\mu \to \infty) = |0 - 0| = 0 \text{ where } H(\mu \to \infty) \to 1. \tag{B.11}$$

**Appendix C.**

There are redundancies in the equations of constraint that must be eliminated before iterative determination of the Lagrange Multipliers. The redundancies can be formulated recursively. For example, in the $MEP[T]$ in Section 2.1 with its three RVs, the lower order marginals are connected to the higher-order ones via the chain:

$$p(123) \to p(12), p(23), p(13); \ p(12) \to p(1); \ p(23) \to p(2); \ p(13) \to p(3). \tag{C.1}$$

For the GMEP in Section 2.2, the redundancies generalize to the pair probabilities satisfying the row-based (the last row is fixed) conditions:

$$p_{t,t'}(i_T, i_{t'}) = p_{t'}(i_{t'}) - \sum_{t=1}^{T-1} p_{t,t'}(i_t, i_{t'}) (1 \leq t < t' \leq T). \tag{C.2}$$

Or they can be formulated as column-based (the last column is fixed)

$$p_{t,t'}(i_t, i_T) = p_t(i_t) - \sum_{t'=1}^{T-1} p_{t,t'}(i_t, i_{t'}) (1 \leq t < t' \leq T). \tag{C.3}$$